\documentclass{optica-article}

\journal{opticajournal} 

\articletype{Research Article}

\usepackage{lineno}

\usepackage{amsmath,amsfonts}
\usepackage{algorithmic}
\usepackage[linesnumbered,ruled,vlined]{algorithm2e}
\usepackage{array}
\usepackage[caption=false,font=normalsize,labelfont=sf,textfont=sf]{subfig}
\usepackage{textcomp}
\usepackage{stfloats}
\usepackage{url}
\usepackage{verbatim}
\usepackage{graphicx}
\usepackage{cite}
\usepackage{epsfig}
\usepackage{xspace}
\usepackage{blindtext}
\usepackage{amsfonts}
\usepackage{algorithmic}
\usepackage{textcomp}
\usepackage{multirow}
\usepackage{adjustbox}
\usepackage{etoolbox}
\usepackage{dsfont}
\usepackage{float}
\usepackage{lipsum}
\usepackage{stfloats}
\usepackage{multicol}
\usepackage{bm}
\usepackage{array}
\usepackage{tabulary}
\usepackage{paralist}
\usepackage{enumerate}
\usepackage{bbding}
\usepackage{pifont}
\usepackage{colortbl}
\usepackage{diagbox}
\usepackage{wrapfig}
\usepackage[normalem]{ulem}
\usepackage{soul}
\usepackage{xcolor}
\sethlcolor{red}

\definecolor{tableHeadGray}{gray}{.9}
\definecolor{tableSubHeadGray}{gray}{0.95}
\newcommand{\PAR}[1]{\noindent{\bf #1}}

\usepackage{color}
\usepackage{booktabs}
\usepackage{placeins}

\makeatletter
\DeclareRobustCommand\onedot{\futurelet\@let@token\@onedot}
\def\@onedot{\ifx\@let@token.\else.\null\fi\xspace}
\def\eg{\emph{e.g}\onedot}

\def\etal{\emph{et al}\onedot}
\makeatother

\begin{document}
\title{SGE: Structured Light System Based on Gray Code with an Event Camera}

\author{Xingyu Lu, Lei Sun\authormark{*}, Diyang Gu, and Kaiwei Wang\authormark{*}}

\address{The College of Optical Science and 
 Engineering, Zhejiang University, 310027 Hangzhou, China
 
 The National Research Center for Optical Instrumentation, Zhejiang University, 310027 Hangzhou, China}

\email{\authormark{*}leo\_sun@zju.edu.cn; wangkaiwei@zju.edu.cn} 


\begin{abstract*} 
Fast and accurate depth sensing has long been a significant research challenge. 
Event camera, as a device that quickly responds to intensity changes, provides a new solution for structured light (SL) systems.
In this paper, we introduce Gray code into event-based SL systems for the first time.
Our setup includes an event camera and a Digital Light Processing (DLP) projector, enabling depth estimation through high-speed projection and decoding of Gray code patterns.
By employing Gray code for point matching in event-based SL system, our method is immune to timestamp noise, realizing high-speed depth estimation without loss of accuracy and spatial resolution. 
The binary nature of events and Gray code minimizes data redundancy, enabling us to fully utilize sensor bandwidth at 100\%. 
Experimental results show that our approach achieves accuracy comparable to state-of-the-art scanning methods while surpassing them in data acquisition speed (up to 41 times improvement) without sacrificing accuracy and spatial resolution. 
Our proposed approach offers a highly promising solution for ultra-fast, real-time, and high-precision dense depth estimation.

\end{abstract*}

\section{Introduction}
Depth estimation plays a pivotal role in various fields\cite{heim2020analysis, xu2020status, fabian2022systematic}, including machine vision and robotics, with applications spanning 3D modeling, autonomous driving, industrial inspection, and more. As a depth estimation method based on the principle of triangulation, the structured light system (SL system) projects a known pattern onto the target and estimates depth based on the pattern spatially modulated by the target\cite{besl1988active, salvi2010state}. The SL system is favored for its high accuracy, dense depth estimation capability, insensitivity to target texture features, and reasonable cost, rendering it indispensable in domains such as 3D scanning and augmented reality\cite{zhang2021high}. Presently, the SL system boasts numerous mature commercial applications, underscoring its widespread adoption and effectiveness.

However, SL systems employing traditional frame-based cameras are restricted to \textit{\textbf{low frequency}} (usually at a frame rate of tens of hertz) due to bandwidth and exposure time constraints of the cameras. While high-speed cameras offer the potential for ultra-fast acquisition\cite{kondo2012development,tao2018high,yin2019high}, their exorbitant cost renders them impractical for many applications.
Moreover, traditional SL systems suffer from \textit{\textbf{data redundancy}}. Images captured by frame-based cameras contain substantial redundant information, particularly in point and line-structured light systems \cite{a26}. This redundancy undermines the acquisition speed and overall efficiency of the system.

Recently, event cameras\cite{brandli2014240,a31,2008A,sun2021mefnet,sun2022event,bao2023improving,sun2023event,bao2024temporal}, also known as Dynamic Vision Sensors (DVS), have emerged as a promising option for SL systems. These bio-inspired devices asynchronously report changes in light intensity at the $\mu s$ level. Due to their capability of solely generating signals for changes in light intensity, event cameras inherently mitigate data redundancy in most SL systems at the hardware level. Additionally, the high dynamic range (HDR) of event cameras allows them to operate effectively under diverse illumination conditions. Consequently, event cameras offer an alternative solution for high-speed depth estimation in SL systems.

Several studies have already explored its potential for depth estimation. Laser-point-projector-based approaches\cite{a4,a5,a6,fu2023fast} have achieved a scanning speed of 60 Hz and high recovery accuracy. After program engineering, they can perform real-time dense target depth estimation at lower frame rates. However, these methods are limited by the low encoding efficiency, timestamp noise (Latency and Jitter) and low scanning rate of the projector, making them hard to achieve higher-speed depth estimation. On the other hand, works based on the full-frame method can achieve higher speed by using Digital Light Processing (DLP) projectors. Among these methods, current temporal coding approaches \cite{a7, a8, a9} are still influenced by timestamp noise and face challenges in robust temporal decoding. Conversely, the spatial coding method \cite{huang2021high} is a one-shot approach that achieves good results in the planar scene lacking detail, but at the cost of the resolution. The trade-off makes it difficult to capture detailed information and results in edge flickering due to spatial decoding.


Therefore, considering the high-speed response and low redundancy characteristics of event cameras, as well as the high-speed projection characteristics of DLP projectors, we propose our Structured light system based on Gray code with an event camera (SGE), a temporal coding method which only uses Gray code to encode the target, and estimates its depth based on epipolar geometry. 
This method first utilizes temporal coding of Gray code, instead of using timestamp matching for point matching or needing accurate timestamps to get a phase map. In contrast to timestamp matching, our SGE enables faster DLP projection and is not influenced by the noise of the event camera and projector. In comparison with phase map methods (current temporal coding methods), our SGE offers robust temporal decoding, faster processing speed, and is also not affected by noise.
Secondly, the binary nature of Gray code perfectly corresponds to the binary response of event cameras. This allows our method to preserve all encoding information without data redundancy.
Thirdly, we have developed a novel calibration scheme for event-based SL systems. Compared to previous schemes\cite{a11,a12,a13}, our scheme has sub-pixel level accuracy and is convenient to use, and it is also applicable to all event-based SL systems.
Finally, we also propose the Gray code X-map (GX-map) disparity query scheme. This scheme enables us to directly query disparities instead of disparity searching. We achieved a processing speed improvement of around 250 times in our experiment. 
These advantages allow us to achieve a much higher depth estimation speed exceeding 1000 Hz while maintaining comparable accuracy to state-of-the-art methods at a millimeter-level precision, which is within a range of 1/200 of the average depth. Meanwhile, the straightforward yet effective pipeline enables our SGE to achieve real-time processing at approximately 200 Hz in 720 * 1280 resolution.

The contributions of this paper are as follows: 
\begin{itemize}
 \item A structured light system based on Gray code with an event camera system (SGE) is proposed with a depth acquisition speed over 1000 Hz (Up to 41 times improvement). Experiments on both static and high-speed dynamic scenes show the superiority of the proposed SGE.
  
 \item  A sub-pixel level, universally applicable calibration method, and the GX-map disparity query scheme are proposed. The latter improves the processing speed by a factor of around 250. 

 \item An optimized depth estimation pipeline has been put forward, enabling an algorithm capable of real-time processing.



\end{itemize}

\section{Related work}

\PAR{Structured light system.} 
Traditional SL system performs depth estimation based on patterns spatial modulated by the target.
As a field that has been researched for many years, it has yielded many accomplishments. Such as different projection pattern: point\cite{a26,a27,a28}, line\cite{a29}, patterned area\cite{a18,a19}, binary coding\cite{a15,a17,a16}, continuous coding\cite{a23,a25}. And as well as different numbers of patterns, including multiple-shot\cite{a24} and single-shot\cite{a20,a21,a22}. These schemes have their advantages and disadvantages in terms of accuracy, speed, environmental adaptability, etc., thus requiring selection based on specific circumstances\cite{a30,a14}. However, due to the limited frame rate of frame-based cameras, typically only tens of Hz, it is impossible for traditional SL systems to achieve high-speed depth estimation. This limitation prevents their application in dynamic scenes.

\PAR{Event-based depth estimation.} 
Due to advantages\cite{a31} of event cameras, for example, high temporal resolution, they have been gradually used in the field of depth estimation.
Initial approaches to event-based depth estimation primarily relied on feature matching with multi-view stereo techniques~\cite{Rebecq16bmvc}. Subsequently, event-based Simultaneous Localization and Mapping (SLAM) methods~\cite{Rebecq17ral,Rosinol18ral,Zhu17cvpr} gained traction for constructing environmental representations or maps and deriving metric depth information. However, these approaches rely on either the scale provided by available camera poses or auxiliary sensors like inertial measurement units (IMUs) for metric depth recovery.
Gallego et al.~\cite{gallego2018unifying} removed dependencies on other modalities with a contrast maximization framework.
Later, learning-based methods utilized events alone to predict depth with multi-view and egomotion constraints. Ye et al.~\cite{ye2018unsupervised} proposed a self-supervised framework for depth, optical flow, and egomotion prediction based on the principle of the egomotion model, which was further improved by Zhu et al.~\cite{zhu2019unsupervised} with Temporal Reprojection Loss, Stereo Disparity Loss, and other constraints. These methods all utilized multi-view approaches, either stereo cameras or egomotion.
Hidalgo et al.~\cite{hidalgo2020learning} were the first to propose predicting dense depth with a monocular camera using supervised learning. Gehrig et al.~\cite{gehrig2021combining} combined frames and events for more robust depth estimation.
However, learning-based methods still have significant limitations in terms of the accuracy of depth estimation. This heavily rely on high-quality datasets and proper training techniques. Furthermore, 
since the data is generated from slow target motion and need to pass through a slow network, both the data collection and processing speed are relatively slow.

\PAR{Event-based SL system.} 
Event cameras have also gained attention in SL systems. Early work\cite{a3} used a 500 Hz pulsed laser to achieve three-dimensional reconstruction by moving the target. However, this method required a displacement stage for scanning. Matsuda \etal\cite{a4} proposed Motion Contrast 3D (MC3D) replacing the pulsed laser with a laser-point projector and utilizing pixel timestamps for point matching. This approach laid a foundation for subsequent research but suffered from quality degradation under high-speed scanning due to timestamp noise. Muglikar \etal\cite{a5} analyzed this noise and improved MC3D's matching algorithm, enhancing robustness against timestamp noise at a scanning speed of 60 Hz. However, it slowed down processing speed due to increased computational effort. Morgenstern \etal\cite{a6} introduced the X-map method, which used table look-up operations for fast disparity calculation and achieved real-time depth estimation. However, timestamp noise interference remained a challenge. Other works explored DLP projection for faster depth estimation. 
Huang \etal\cite{huang2021high} employs a spatial coding method by using pseudo-random black-and-white dots and realizes a scan rate under 1000 Hz. On the measurement of the simple planar scene, they use the plane fitted by depth points they obtained as ground truth and calculate the root mean square error (RMSE) of measurement points and fitting points as their accuracy metrics, and achieve good results. However, this method further reduces the resolution when the resolution of the event camera is already not high, which makes it difficult to capture detailed information. It also suffers from edge flickering and complex data processing resulting from spatial decoding. 
Current works\cite{a7,a8,a9} based on temporal coding method faced issues of realizing high accuracy and robust decoding. In summary, most existing works struggle to achieve high accuracy, high speed, and robust depth estimation simultaneously but show favorable results only in specific aspects.

\section{Method}
We first introduce the foundation and basic consideration of our SGE method in Sec.~\ref{sec:foundation} and then outline the pipeline in Sec.~\ref{sec:pipeline}. In Sec.~\ref{sub:calibration}, our simple and high-accuracy calibration scheme for event-based SL systems will be proposed. Finally, we detail the depth estimation scheme based on the GX-map query and Time-overlapping strategy in Sec.~\ref{sub:estimation}.

\subsection{Preliminaries and Analysis}
\label{sec:foundation}

\PAR{SL system for depth estimation.}~Traditional SL systems, composed of a frame-based camera and a projector, typically employ triangulation for depth estimation. Similar to common stereo vision systems, SL systems can be conceptualized as artificially projecting "features" onto the target surface using a projector. Due to the typically dense nature of this projection, SL systems can achieve denser depth estimation, surpassing stereo vision, especially in generating highly accurate depth information at short distances\cite{schwarte1999handbook}. The primary distinction among SL schemes lies in the use of different projection patterns. However, these patterns only lead to variations in depth encoding and do not impact the disparity calculation process. Therefore, in structured light depth estimation, after obtaining the depth encoding, stereo rectification is first performed based on the calibration parameters, and then point matching: 

\vspace{-0.5em}
\begin{equation}
P_{pr}(x_{pr}, y_{pr}) = F(P_{cr}(x_{cr}, y_{cr})),
\end{equation}

where $P_{cr}$ denotes an encoded point in the camera's pixel coordinate system, and $P_{pr}$ represents the corresponding point in the projector's pixel coordinate system. $x_{pr}$, $y_{pr}$, $x_{cr}$, $y_{cr}$ represent pixel coordinates in the corresponding pixel coordinate system. And $F$ represents the matching algorithm. Assuming that the disparity direction is along the $X$ direction, the depth can be obtained by:

\vspace{-0.5em}
\begin{equation}
\label{eq:triangular}
Z(P) = \frac{f \cdot b}{x_{cr}-x_{pr}}.
\end{equation}

Here, $f$ denotes the focal length, while $b$ represents the baseline distance. However, SL systems encounter two basic limitations preventing them from high-speed measurement. First, it is restricted by \textit{\textbf{low frame rate}} of the frame-based camera, typically only tens of hertz. Second, the \textit{\textbf{abundant data redundancy}} of images not only takes up sensor bandwidth but also demands additional data processing\cite{a5}.

\PAR{Gray code SL system.}~The Gray code SL system uses patterns with binary stripes for projection\cite{a30}. Thus, each point along the disparity direction has a unique binary code for later point matching. Its binary nature makes the method robust and adaptable. However, this nature also leads to low grayscale information utilization compared to other grayscale-based methods, requiring a large number of patterns for high spatial resolution, which increases data acquisition time. 
But with an event camera, only binary information is obtained from the hardware to speed up data acquisition, making Gray code perfect for the event-based system.
All the above answers the question of why this work combines the Gray code and the event camera.



\PAR{Event generation.}~The recently popular event cameras provide a novel solution for high-speed structured light depth estimation thanks to their high temporal resolution. Each pixel of the event camera reports the intensity changes after perceiving intensity change exceeding the threshold, which we call events ($E=(x, y, \tau, p)$), where $x$, $y$, $\tau$, $p$ represent the pixel coordinates, output timestamp, and polarity, respectively. For an event $e$, it's polarity $p_{e,\tau}$ can be formulated as: 

\vspace{-0.5em}
\begin{equation}
\label{eq:event_polarity}
p_{e,\tau} = 
\begin{cases}
+1, \text{if} \log \left( \frac{\mathcal{I}_{t,e} } {\mathcal{I}_{ (t - \Delta t),e}  } \right) > C_{positive}, \\
-1, \text{if} \log \left( \frac{\mathcal{I}_{t,e} } {\mathcal{I}_{ (t - \Delta t),e}  } \right) < C_{negative},
\end{cases}
\end{equation}
where
\begin{equation}
    \tau = t + \sigma,
    \label{eq:tau}
\end{equation}
$\mathcal{I}$ represents intensity, $C_{positive}$ and $C_{negative}$ represent the threshold for generating positive and negative events, which can be manually adjusted. $t$, $\sigma$, and $\tau$ are the ideal timestamp, the timestamp noise, and the actual output timestamp, respectively.

The previous works\cite{a5,a6,huang2021high,a7,a8,a9} have utilized event-based SL systems to realize depth estimation. However, point-scanning approaches suffer from timestamp noise and low encoding efficiency. The former leads to a decrease in accuracy when scanning speed increases. The latter means more timestamps are needed for depth encoding, as shown in Equ.~\ref{eq:complexity1}, where $n$ is the number of timestamps needed for encoding, and $c$ is the number of columns (disparity direction). The temporal coding has issues such as not robust temporal decoding and is still affected by timestamp noise. Meanwhile, the spatial coding has other problems like low spatial resolution and edge flickering.

\begin{equation}
\label{eq:complexity1}
n = c,
\end{equation}

\begin{equation}
\label{eq:complexity2}
n = \log_2(c),
\end{equation}

Therefore, for the first time, we propose to use Gray code as a depth encoding pattern for depth estimation in event-based SL systems. Firstly, compared to point scanning methods, by using Gray code, we achieve \textit{\textbf{robust}} temporal decoding and \textit{\textbf{higher encoding efficiency}} while maintaining the absence of redundant information, as shown in Equ.~\ref{eq:complexity2}. Secondly, our approach is \textit{\textbf{immune}} to timestamp noise because we only need timestamps for data segmentation. Thirdly, compared with the spatial coding methods, which sacrifice spatial resolution for acquisition speed, our approach \textit{\textbf{remains the origin spatial resolution}}. These illustrate the capacity of our SGE to achieve high speed, high-precision depth estimation, and high-resolution details.

\subsection{Overall Pipeline}
\label{sec:pipeline}

\begin{figure*}[t]
    \centering
    \includegraphics[width=0.95\textwidth]{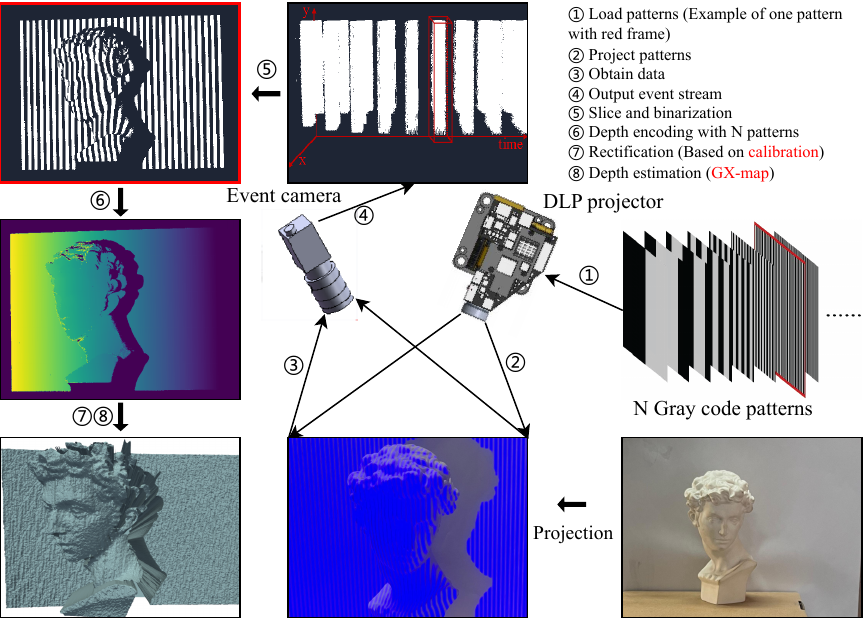}
    \vspace{-0.5em}
    \caption{\textbf{The pipeline of the proposed SGE.} Patterns are loaded and cast onto a target, and then captured by an event camera in the event stream type continuously. Then binarized images are obtained by conducting step (5). When a set of images is gathered, steps (6-8) are conducted to estimate depth.}
    \label{fig:overall}
    \vspace{-1.5em}
\end{figure*}

Our SGE comprises a DLP projector and an event camera, as depicted in Fig.~\ref{fig:overall}. Before deployment, we first calibrate the system (Sec.~\ref{sub:calibration}) using our proposed Simple Event-based SL system Calibration (SEC). Then the projector independently and continuously projects Gray code onto the target. Thanks to DLP's expertise in binary projection, our system can achieve an exceptionally rapid projection speed of up to 2487 Hz, which is only limited by our hardware. Patterns modulated by the target depth are captured by the event camera from another view.
As the data from the event camera is in the form of an event stream, segmentation according to each projection and subsequent conversion into image form are required.
After acquiring a whole set of patterns, we use the traditional Gray code algorithms\cite{a30,posdamer1982surface} to derive the depth-encoded image. Finally, leveraging the system's calibrated parameters and GX-map, we determine the depth in Sec.~\ref{sub:GX-map}. For dynamic scenes, we propose the Time-overlapping strategy to achieve higher data utilization and depth estimation speed (Sec.~\ref{sub:Time-overlapping}). We also conduct experiments on the low computational cost of our proposed SGE in Sec.~\ref{sec:realtime} to show the real-time processing capability. Finally, Alg.~\ref{Alg:alg} shows the pseudo-code of the proposed SGE.

\begin{figure}[t]

    \centering
    \includegraphics[width=0.7\textwidth]{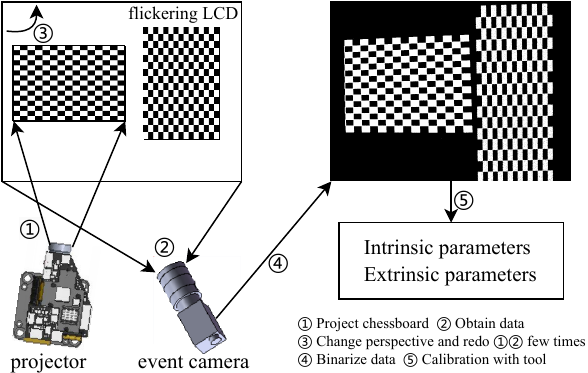}
    \caption{\textbf{The calibration pipeline of our SEC.} The event camera captures a projected checkerboard and a flickering checkerboard on the same plane from various perspectives. Then binarizing them into images and using the Gabriel calibration tool to calibrate SL systems.}
    \label{fig:cal_process}

\end{figure}

\subsection{Simple Event-based SL System Calibration}
\label{sub:calibration}


Previous work\cite{a5} converts the event to images with E2VID~\cite{rebecq2019high} and utilizes calibration tool~\cite{a11} for rough extrinsic parameters calibration. Alternatively, some\cite{a6} adopt a step-by-step strategy and separately calibrate the camera's intrinsic parameters. However, these methods (\textit{i}) necessitate dual calibration procedures and a complex calibration process, rendering them cumbersome to use. And (\textit{ii}) images generated using E2VID may struggle to maintain sub-pixel level authenticity, while the step-by-step strategy introduces significant artificial accuracy loss. While some works have achieved high-precision calibration \cite{a12}, they are limited to laser-point projectors or event cameras with gray mode.

\begin{figure}[h]
    \centering
    \includegraphics[width=0.7\textwidth]{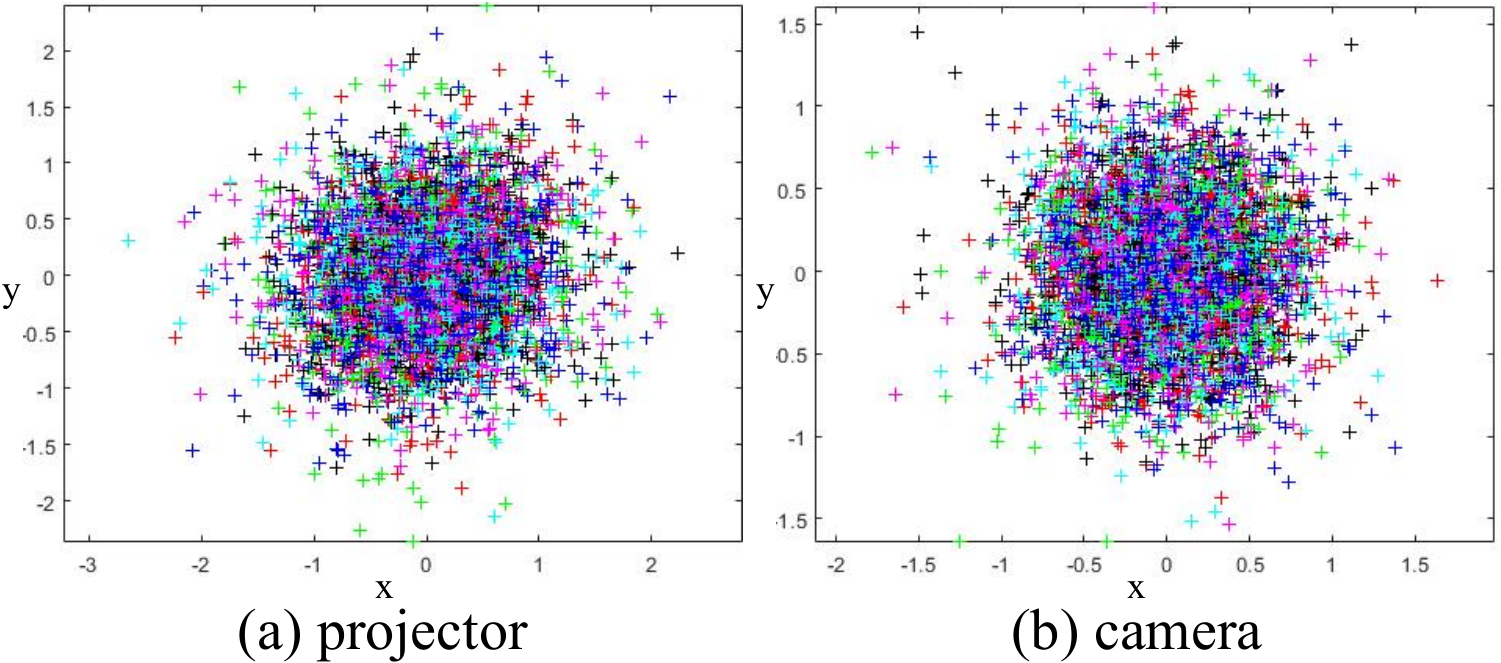}
    \caption{\textbf{The calibration reprojection error of the projector and camera}. The horizontal and vertical axes are measured in pixels. The average reprojection error and rough distribution range of [0.59, 0.54], [-2, 2] for the projector and [0.40, 0.40], [-1.5,1.5] for the event camera.}
    \label{fig:reproj}
\end{figure}

Therefore, based on Gabriel's calibration tool\cite{falcao2008plane}, we propose a novel, sub-pixel level, simple, and versatile event-based SL system calibration scheme SEC, as shown in the Fig.~\ref{fig:cal_process}. 
Initially, we affix an LCD screen presenting a flickering chessboard onto a planar surface and project another chessboard pattern onto the other side. Following the acquisition of multiple datasets from various perspectives, the data is directly binarized and subjected to calibration procedures using the standard calibration tool. Our approach facilitates simultaneous calibration of all parameters through straightforward operations, significantly mitigating the complexity associated with calibration processes. Furthermore, our method attains high calibration precision, demonstrating the capability to achieve sub-pixel level reprojection errors.
As shown in Fig.~\ref{fig:reproj}, we achieve the average reprojection error and rough distribution range of [0.59, 0.54], [-2, 2] for the projector and [0.40, 0.40], [-1.5,1.5] for the event camera in the experiment with 23 sets of data collected from different poses. Moreover, our SEC is universally applicable to all event-based SL systems.

\begin{algorithm*}[t]

  \SetAlgoLined
  \SetKwInOut{Input}{Input}
  \SetKwInOut{Output}{Output}
  
  \Input{Depth-encoded image of projector: $D_{p}$  \\
   Event data: $E$    \\
   Calibration(intrinsic and extrinsic parameters): $C$  \\    
  }
  \Output{Depth map: $D$}
  
  \BlankLine
  \tcc{Initialization}
  \tcp{Obtain the rectified depth-encoded image of projector}
   $D_{pr}$ $\leftarrow$ ($D_{p}$, $C$) \;
  \tcp{Obtain GX-map according to Equ.~\ref{eq:GX}}
   $GX()$ $\leftarrow$ $D_{pr}$ \;
  
  \BlankLine
  \tcp{Time-overlapping strategy}
  \tcc{Main loop}
  \tcp{Event data is not over}
  \While{E $\neq$ None}
  {
    \tcp{Sequential acquisition of Gray code slice data and binarization (Step (a))}
    $G_{x}$ $\leftarrow$ $E$ \\
    \tcp{A complete set of Gray code is obtained}
    \If{$n(G)$ $=$ $N$ }{
    \tcp{Depth-encoded image of camera calculation (Step (b))}
    $D_{c}$ $\leftarrow$ $G$ \\
    \tcp{Rectification (Step (c))}
    $D_{cr}$ $\leftarrow$ ($D_{c}$, $C$)\\
    \tcp{Disparity query according to Equ.~\ref{eq:queryx} (Step (d))}
    $dis$ $\leftarrow$ ($D_{cr}$,$GX()$)\\
    \tcp{Depth map calculation according to Equ.~\ref{eq:triangular}}
    $D$ $\leftarrow$ ($dis$, $C$)\\
    }
    \BlankLine
  }
  
  \BlankLine
  
  \caption{Depth estimation by SGE method}
\label{Alg:alg}

\end{algorithm*}

\subsection{Depth Estimation}
\label{sub:estimation}
\begin{figure}[t]
    \centering
    \includegraphics[width=0.8\textwidth]{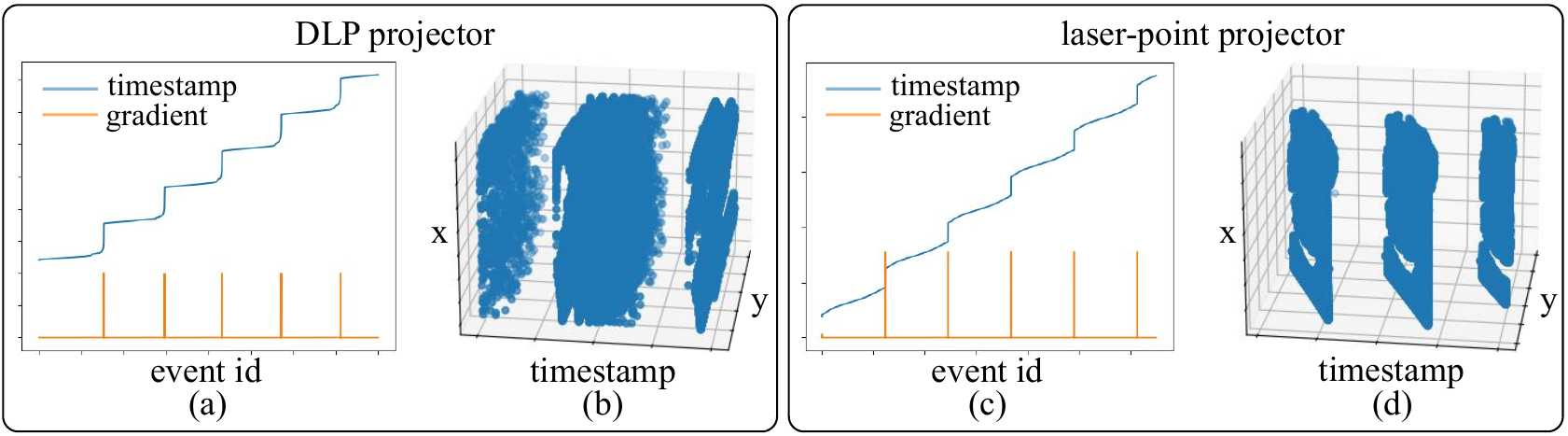}
    \caption{\textbf{The projection characteristics of DLP projectors and laser-point projectors.} 
    (a) and (c) show the timestamp and time gradient variations with event id (sorted by time). (b) and (d) present the scatter plot of the events during the projection. DLP employs multi-point scanning, while laser-point uses raster scanning with non-linear characteristics. Both include dark intervals between projections.}
    \label{fig:projector_cha}
\end{figure}

\begin{figure}[t]
    \centering
    \includegraphics[width=0.8\textwidth]{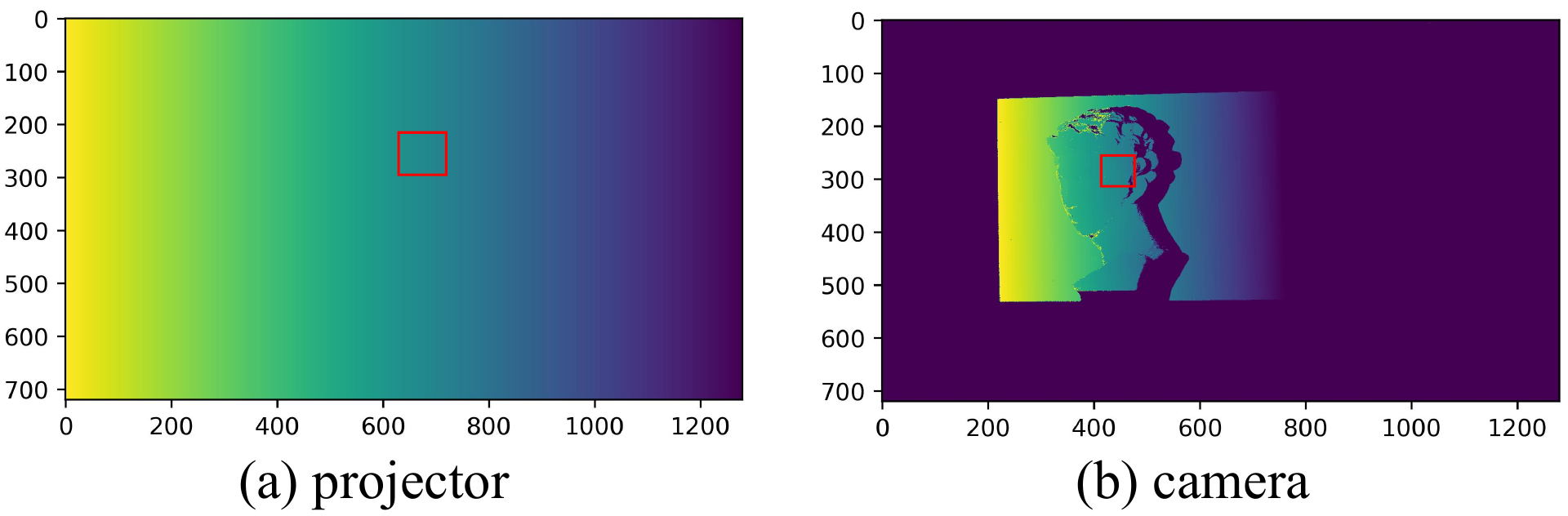}
    \vspace{-0.5em}
    \caption{\textbf{Depth-encoded images}. Different colors represent different depth encodings. The red box indicates the approximate matching position. 
    }
    \label{fig:timemap}
\end{figure}

\PAR{Depth-encoded Image Acquisition.}
As mentioned in Sec.~\ref{sec:pipeline}, the continuous projection data captured by the event camera manifests in the form of an event stream. Consequently, the initial step involves segmenting the data based on each projection. The presence of projection dark time intervals, as illustrated in Fig.~\ref{fig:projector_cha}, enables straightforward identification of time windows within the event stream devoid of event occurrences, which serve as natural boundaries for segmentation. Subsequently, the segmented slices are binarized into Gray code images, following which traditional Gray-coded depth estimation algorithms\cite{a30} are employed to derive depth-encoded image of the target, as demonstrated in Fig.~\ref{fig:timemap}.

\begin{figure}[t]
    \centering
    \includegraphics[width=0.8\textwidth]{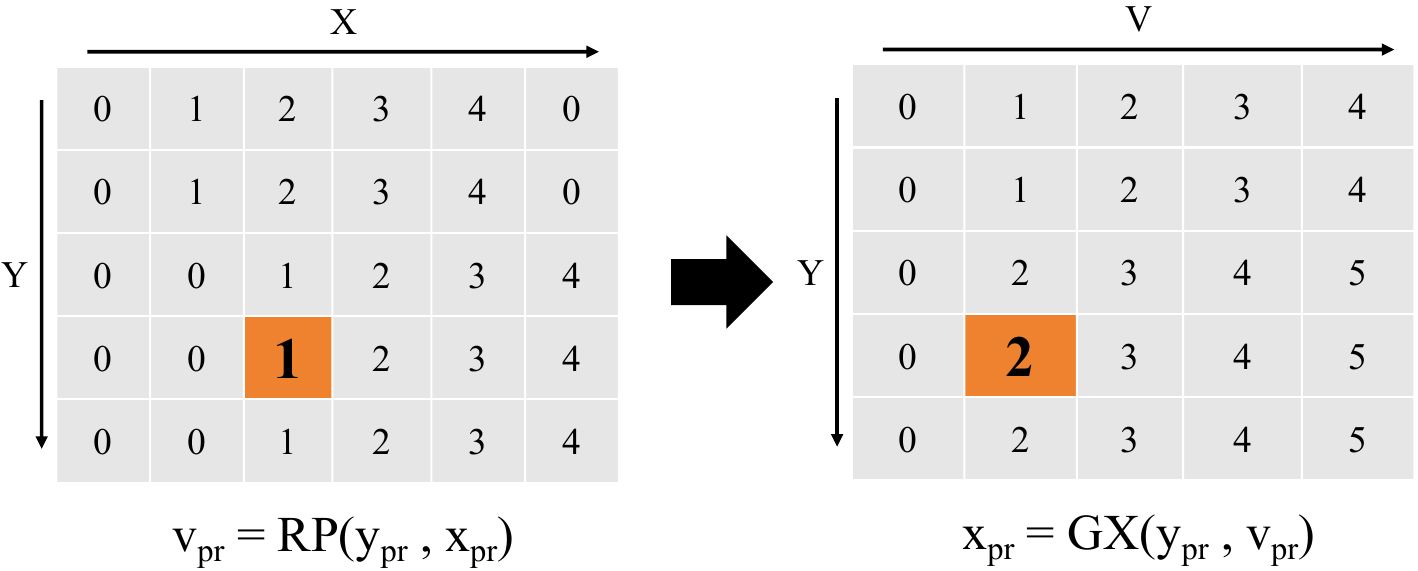}
    \caption{\textbf{\textbf{Schematic diagram of GX-map generation.}} Transform the rectified depth-encoded image of the projector that queries encoding values based on Y and X coordinate values into the GX-map that queries X coordinate values based on Y coordinate and encoding values. For example, the orange region are changed from “1 = RP(3,2)” to “2 = GX(3,1)”}
    \label{fig:GX-map}
\end{figure}

\PAR{GX-map Disparity Query.}
\label{sub:GX-map}
Following the acquisition of depth-encoded images, point matching is essential to extract depth information.
Due to the \textit{\textbf{high time-consuming}} nature of epipolar line search, we propose the GX-map disparity query scheme tailored for Gray code. 
Upon conducting stereo rectification to obtain rectified depth-encoded images of the projector and camera, we first store the x coordinates $x_{pr}$ of the projector's rectified depth-encoded image over y coordinates $y_{pr}$ and the encoding values $v_{pr}$ to generate GX-map. As illustrated in Fig.~\ref{fig:GX-map}, the left side is the rectified depth-encoded image of the projector, the right side is the corresponding GX-map, and the orange area represents the same data before and after conversion. Then, due to the fact that the matched points pair in rectified depth-encoded images of the projector and camera have the same y coordinates and encoding values, we can directly query the corresponding disparity $d$ using the rectified camera's depth-encoded image. These are formulated as:


\vspace{-1em}
\begin{eqnarray}
\label{eq:GX}
v_{pr} = RP(y_{pr},x_{pr}) \rightarrow x_{pr} = GX(y_{pr},v_{pr}),
\end{eqnarray}

\vspace{-2em}
\begin{eqnarray}
\label{eq:queryx}
d = x_{cr} - GX(y_{cr},v_{cr}),
\end{eqnarray}

where $x_{cr}$, $y_{cr}$ and $v_{cr}$ represent the coordinates and encoding value of the camera's rectified depth-encoded image, respectively. $RP()$ and $GX()$ represent functions that query based on the projector's rectified depth-encoded image and GX-map. Given that the rectified depth-encoded image of the projector remains unchanged under identical intrinsic and extrinsic parameters, the GX-map generation operation only needs to be executed once and can then facilitate multiple queries. Through GX-map querying, we have attained a \textit{\textbf{200 times enhancement}} in computational speed compared to traditional epipolar line search on the same processor, making high-speed real-time depth estimation possible.

\begin{figure}[t]
    \centering
    \includegraphics[width=0.8\textwidth]{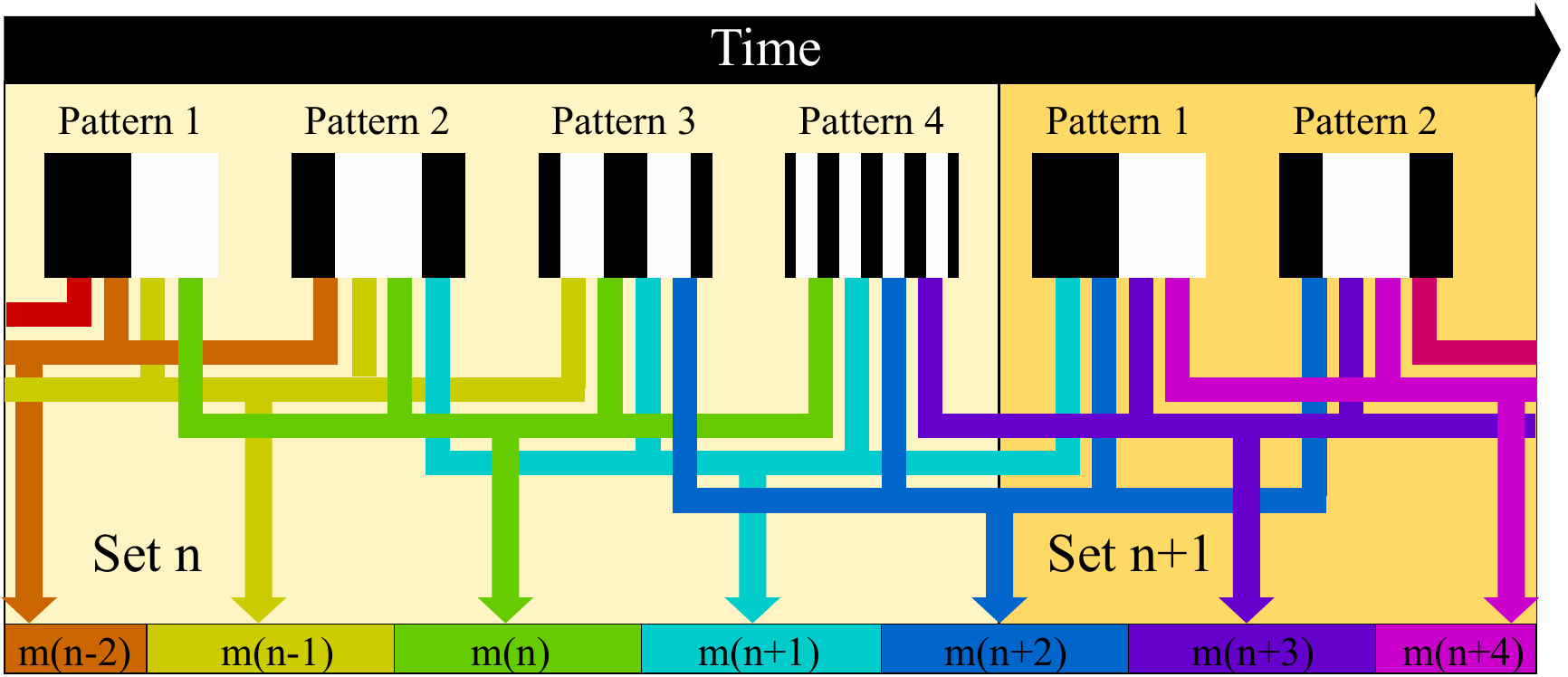}
    \caption{\textbf{Time-overlapping strategy of Gray Code}. We take a 4-bit Gray code as an example, each consecutive 4 images can generate a corresponding depth-encoded image, greatly improving data utilization and depth estimation speed.}
    \label{fig:4gray}
\end{figure}

\PAR{Time-overlapping.}
\label{sub:Time-overlapping}
For dynamic scenes, since a set of Gray codes starting with any serial number can compose a depth-encoded image, therefore, similar to\cite{a10}, we propose a Time-overlapping strategy to maximize data utilization, as illustrated in Fig.~\ref{fig:4gray}.

For depth estimation using N-bit Gray code, N projection patterns are required to compute depth. After capturing the initial set of binarized Gray codes, each subsequent Gray code can be combined with the previous images to calculate a new depth-encoded image. This means each projection corresponds to a new depth estimation, further improving the speed of the depth estimation by N times for N-bit Gray code.

\section{Experiments}


\subsection{Settings}
\label{sec:detail}
\begin{figure}
    \centering
    \includegraphics[width=0.7\textwidth]{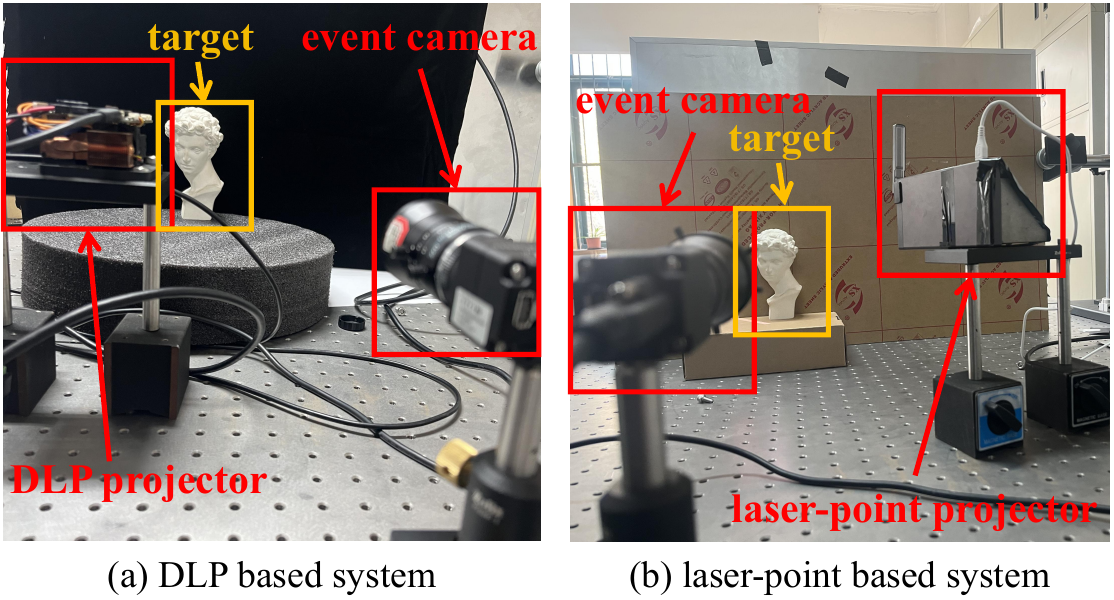}
    \caption{\textbf{The hardware setup}, involving two projectors with different field of view (FoV) and resolutions. Calibration factors are taken into account, resulting in differences in the structure of the two setups.}
    \label{fig:hardware}
\end{figure}

\PAR{Hardware.}
In our experiment, we employ two hardware setups, both securely mounted on the optical platform to maintain a consistent position, as shown in  Fig.~\ref{fig:hardware}. Both setups utilize the Prophesee EVK4 HD camera, with a resolution of 1280$\times$720 and a pixel size of 4.86$\mu m$. 

In the experiment of static scenes depth estimation, we use the Sony Mobile projector MP-CL1A with a resolution of 1920$\times$720 pixels and a fixed projection speed of 60 Hz. It is a laser-point projector working in a raster scanning pattern, which projects clear images at any distance. However, it suffers from laser point diffusion while projection.

In the rest of the experiments, we adopt the DLP projector OPR305185 from Bopixel with the resolution of 1280$\times$720. The fastest projection speed of 1-bit patterns of the projector is 402$\mu s$ per slide, which is well above the temporal resolution of the event camera. More detailed information about the hardware and calibrated data of two systems are provided in the Supplemental document and Datasets 1, 2.

\PAR{Compared methods.}
We choose the state-of-the-art point scanning methods: MC3D\cite{a4}, the initial and optimized ESL\cite{a5}, and the X-map\cite{a6} in the scanning speed of 60 Hz as our comparing methods. We average the result from MC3D\cite{a4} within one second following~\cite{a5}. It is worth mentioning that our calibrated parameters are adopted by other methods.

\PAR{Accuracy assessment.}
In this task we want to show that our SGE achieves comparable accuracy to SOTA methods, therefore we compare our method with each of them to show the difference between them to verify the relative accuracy of our method instead of choosing one of SOTA methods as ground truth (GT). An available GT is difficult to obtain because it is hard to find a method that can generate dense depth of natural scenes above millimeter level accuracy, which is much more exact than all SOTA methods. Therefore previous works use similar different ``reference'' GT\cite{a5,a6}, use different precision metrics\cite{a9,huang2021high} or just qualitative comparison\cite{a4,a7}. Because our SGE and SOTA methods boast similar accuracy level, choosing one of them as GT may more exhibit the disparities in data types (point scanning vs. full-frame) and processing methods (time stamp matching vs. encoding value matching) rather than accuracy. Therefore we just compare the similarity between our scheme and SOTA methods, which is sufficient to demonstrate our comparable accuracy.

\PAR{Metrics.}
The root mean square error (RMSE) and the fill-rate (FR) are adopted as our evaluation metrics. 
To minimize the influence of data differences, like slight differences in edge positions, we only use solid regions in both methods to calculate RMSE. We define solid regions as where both methods give a reliable depth estimation, which means the difference between them is within the threshold. The FR measures the ratio of points below the depth error threshold in SGE to measured points in the compared method. The depth error threshold is 1\% of the average scene depth. 

\subsection{Static Scenes}
\label{sec:static}

\begin{figure*}[t]
    \centering
    \includegraphics[width=0.95\textwidth]{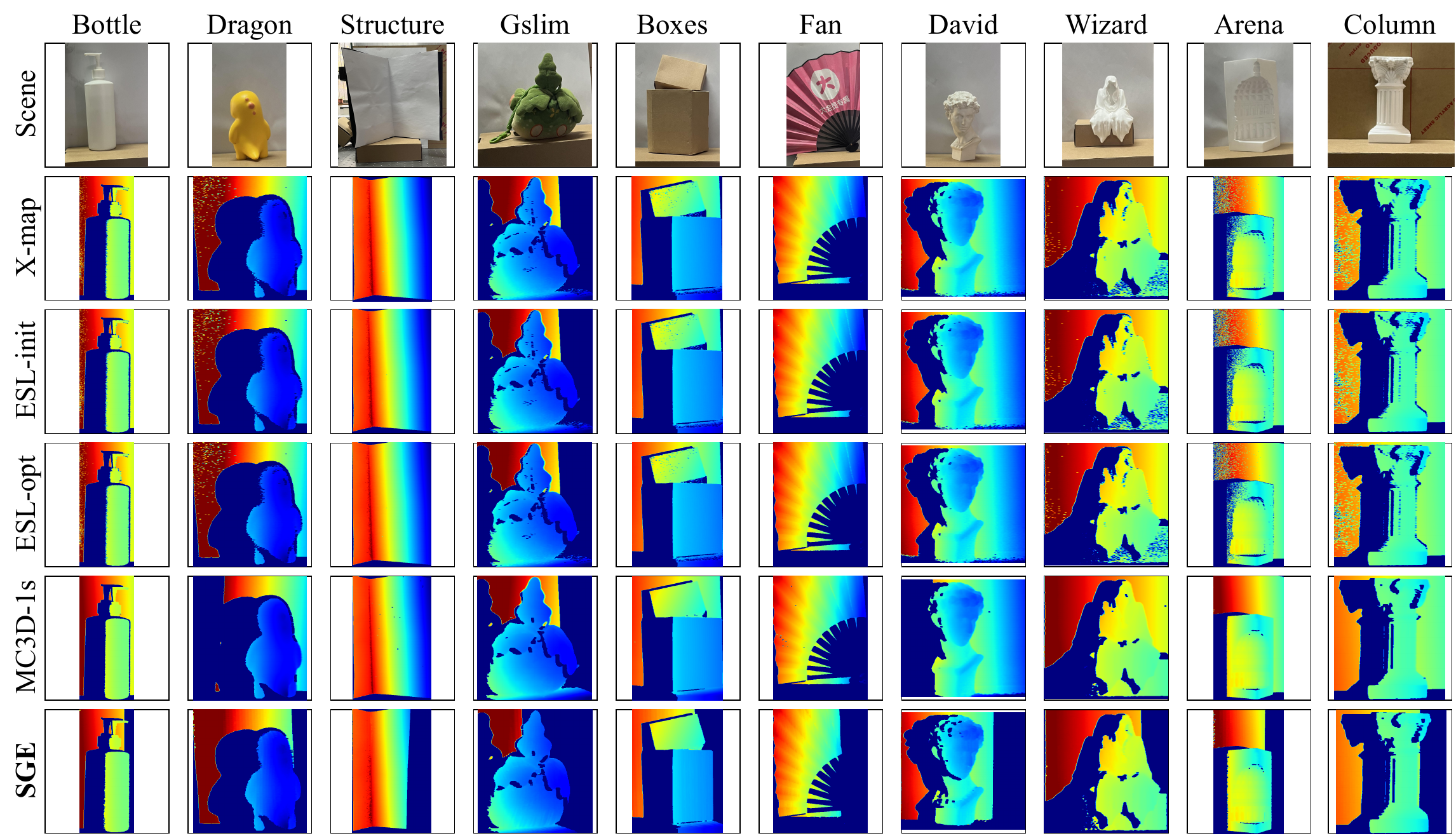}
    \caption{\textbf{Qualitative results of different methods on static scenes with the laser-point based system.} Our SGE not only shows performance consistent with the the-state-of-art schemes but also presents more solid results in concave regions and less noise in overexposed and underexposed locations.}
    \label{fig:static}
\end{figure*}

In the experiment, we captured multiple scenes with varying complexity, including highly detailed gypsum statues, surfaces with fewer details, and everyday scenes to show the versatility of our solution. 
To minimize variables, we adopt the laser-point based system (as shown in  Fig.~\ref{fig:hardware} (b)) to eliminate device differences due to the inability of the point scanning schemes to utilize DLP projectors. We also keep the projection light intensity, ambient light intensity, camera threshold, and shooting position constant while capturing two types of data. During the experiment, we employ the close operation as pre-processing in our SGE to fill in holes.
Since the projector's 720 horizontal resolution cannot be fully utilized by Gray code encoding, we use only 512 columns for projection, which results in the missing background on the right side in our SGE.

Fig.~\ref{fig:static} presents the qualitative depth estimation results. 
During this experiment, we only use mean filtering to reduce discontinuity in all methods as post-processing. 
It can be observed that for point scanning methods, due to timestamp noise and laser point diffusion, there is more error present at the concave regions, \eg Structure and Boxes, and overexposed and underexposed locations, \eg Wizard, Arena, and Column. This is improved in the MC3D scheme by averaging. However, our SGE demonstrates \textit{\textbf{strong resistance}} to timestamp noise and laser point diffusion and performs accurate depth estimation in any case.

\begin{table*}[t]
\caption{\textbf{Quantitative results of the difference between our SGE and other methods in static scenes.} RMSE and FR are adopted as evaluation protocols. Our method achieves an RMSE of less than 2.34mm, which is less than 1/200 of the average depth, and an FR over 0.91 compared to other methods.}
        \label{tab:acq}
	\centering
        \small
        \resizebox{1.0\textwidth}{!}{
                \setlength{\tabcolsep}{6pt}
                \renewcommand\arraystretch{1.0}
    	    \begin{tabular}{rcccccccccccccccccccc}
    		\bottomrule[0.15em]
    		\multirow{1}*{Scene} & \multicolumn{2}{c}{Bottle} & \multicolumn{2}{c}{Dragon} 
                & \multicolumn{2}{c}{Structure} & \multicolumn{2}{c}{Gslim}
    		& \multicolumn{2}{c}{Boxes}     & \multicolumn{2}{c}{Fan}
                & \multicolumn{2}{c}{David}    & \multicolumn{2}{c}{Wizard}
                & \multicolumn{2}{c}{Arena}     & \multicolumn{2}{c}{Column}
                \\
                
    		\multirow{1}*{Mean Depth} & \multicolumn{2}{c}{320mm} & \multicolumn{2}{c}{410mm}
                & \multicolumn{2}{c}{530mm} & \multicolumn{2}{c}{360mm}
                & \multicolumn{2}{c}{510mm} & \multicolumn{2}{c}{530mm}
                & \multicolumn{2}{c}{450mm} & \multicolumn{2}{c}{400mm}
                & \multicolumn{2}{c}{430mm} & \multicolumn{2}{c}{320mm}
                \\
                
                \cmidrule(lr){2-3}\cmidrule(lr){4-5}\cmidrule(lr){6-7}\cmidrule(lr){8-9}
                 \cmidrule(lr){10-11}  \cmidrule(lr){12-13} \cmidrule(lr){14-15}\cmidrule(lr){16-17}
                 \cmidrule(lr){18-19} \cmidrule(lr){20-21}
                 
                Metrics & FR ${\uparrow}$ & RMSE ${\downarrow}$ & FR ${\uparrow}$ & RMSE ${\downarrow}$ & FR ${\uparrow}$ & RMSE ${\downarrow}$ & FR ${\uparrow}$ & RMSE ${\downarrow}$ & FR ${\uparrow}$ & RMSE ${\downarrow}$ & FR ${\uparrow}$ & RMSE ${\downarrow}$& FR ${\uparrow}$ & RMSE ${\downarrow}$ & FR ${\uparrow}$ & RMSE ${\downarrow}$& FR ${\uparrow}$ & RMSE ${\downarrow}$& FR ${\uparrow}$ & RMSE ${\downarrow}$\\
    		\midrule
    		MC3D-$1s$  & 0.99 & 0.70 & 0.98 & 1.00 & 0.96 & 1.68 
                         & 0.95 & 1.44 & 0.97 & 1.84 & 0.99 & 1.68
                         & 0.94 & 1.09 & 0.98 & 1.00 & 0.96 & 1.06
                         & 0.98 & 0.63\\

    		ESL   & 0.99 & 0.62 & 0.99 & 0.87 & \textbf{0.91} & \textbf{2.34}  
                      & 0.94 & 1.11 & 0.96 & 1.52 & 0.97 & 1.58
                      & 0.95 & 1.20 & 0.98 & 0.78 & 0.96 & 1.15
                      & 0.99 & 0.53\\
                      
                ESL-opt   & 0.99 & 0.62 & 0.98 & 0.95 & 0.94 & 1.94 
                          & 0.94 & 1.20 & 0.95 & 1.76 & 0.97 & 1.63
                          & 0.95 & 1.10 & 0.98 & 0.97 & 0.96 & 1.10
                          & 0.99 & 0.55\\
                          
    		  Xmap  & 0.99 & 0.62 & 0.99 & 0.90 & \textbf{0.91} & \textbf{2.34}
                      & 0.94 & 1.12 & 0.96 & 1.52 & 0.97 & 1.61
                      & 0.95 & 1.22 & 0.98 & 0.79 & 0.96 & 1.16
                      & 0.99 & 0.54\\
    		\bottomrule
    	\end{tabular}
        }
        
\end{table*}

The quantitative results are shown in Tab.~\ref{tab:acq}. Our method achieves an RMSE of \textit{\textbf{less than 2.34mm}}, which is \textit{\textbf{less than 1/200}} of the average depth, and an FR \textit{\textbf{over 0.91}} compared to other state-of-the-art methods. This indicates that our method achieves high accuracy comparable to the state-of-the-art with precision at the \textit{\textbf{millimeter level}}. 
During the comparison, considering the data gap and the fairness of the comparison, 
we utilize simple median filtering in the 
post-processing stage to optimize the results for all the methods. It is worth mentioning that our method can be further improved if additional off-line optimization algorithms are adopted at the cost of longer processing time, like those used in ESL.

\subsection{Results Under Different Scanning Speed}
\label{sec:speed}

\begin{figure*}[t]
    \centering
    \includegraphics[width=0.95\textwidth]{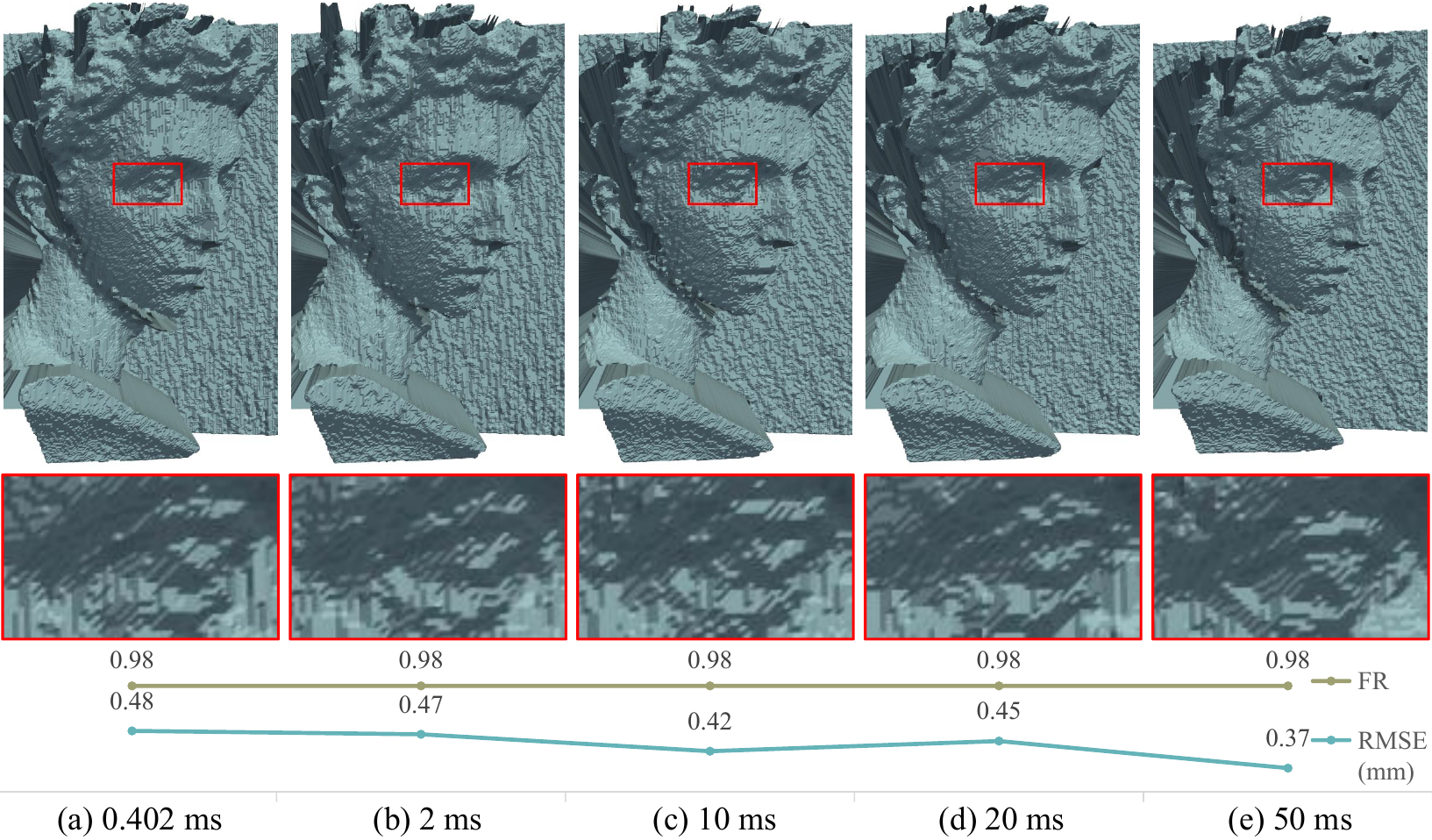}
    \caption{\textbf{Depth estimation of the sculpture at different scanning speeds.} Figures (a)-(e) show the results at different scanning speeds with a quantitative line chart. Averaged result of 50$ms$ is adopted as a reference, therefore the quantitative result of 50$ms$ can also be considered as the repeatability accuracy of our scheme. Our method exhibits consistent performance, both in overall shape and in fine details, across different speeds.}
    \label{fig:HZ}
\end{figure*}


 To demonstrate the accuracy of our SGE under high-scanning-speed conditions, we evaluate its performance at different scanning speeds using the DLP-based system, as shown in Fig.~\ref{fig:HZ}. During the experiments, since the accuracy in low scanning speed has been guaranteed in Sec.~\ref{sec:static}, we use the average results of 29 depth estimations conducted under a scanning speed of 50$ms$ per slide as the reference and then calculate the average metrics between the 29 single result at each scanning speed and the reference. It is not difficult to find that, under different scanning speeds, our SGE produces the \textit{\textbf{approximate performance}} both in overall and details. This means our method can transition from low-speed to high-speed \textit{\textbf{without precision loss}}, proving our capacity of \textit{\textbf{high-speed}} and \textit{\textbf{high-accuracy}} depth estimation and making high-speed and high-accuracy \textit{\textbf{dynamic scenes}} depth estimation possible. 
 
 It is worth mentioning that the highest estimation speed of 402$\mu s$ per slide is not restricted by our method but by the scanning speed of the adopted projector. Therefore using more recent DLP projectors with higher scanning speed can further improve our depth estimation speed.

\subsection{Dynamic Scenes}
\label{sec:dynamic}

\begin{figure*}[t]
    \centering
    \includegraphics[width=0.95\textwidth]{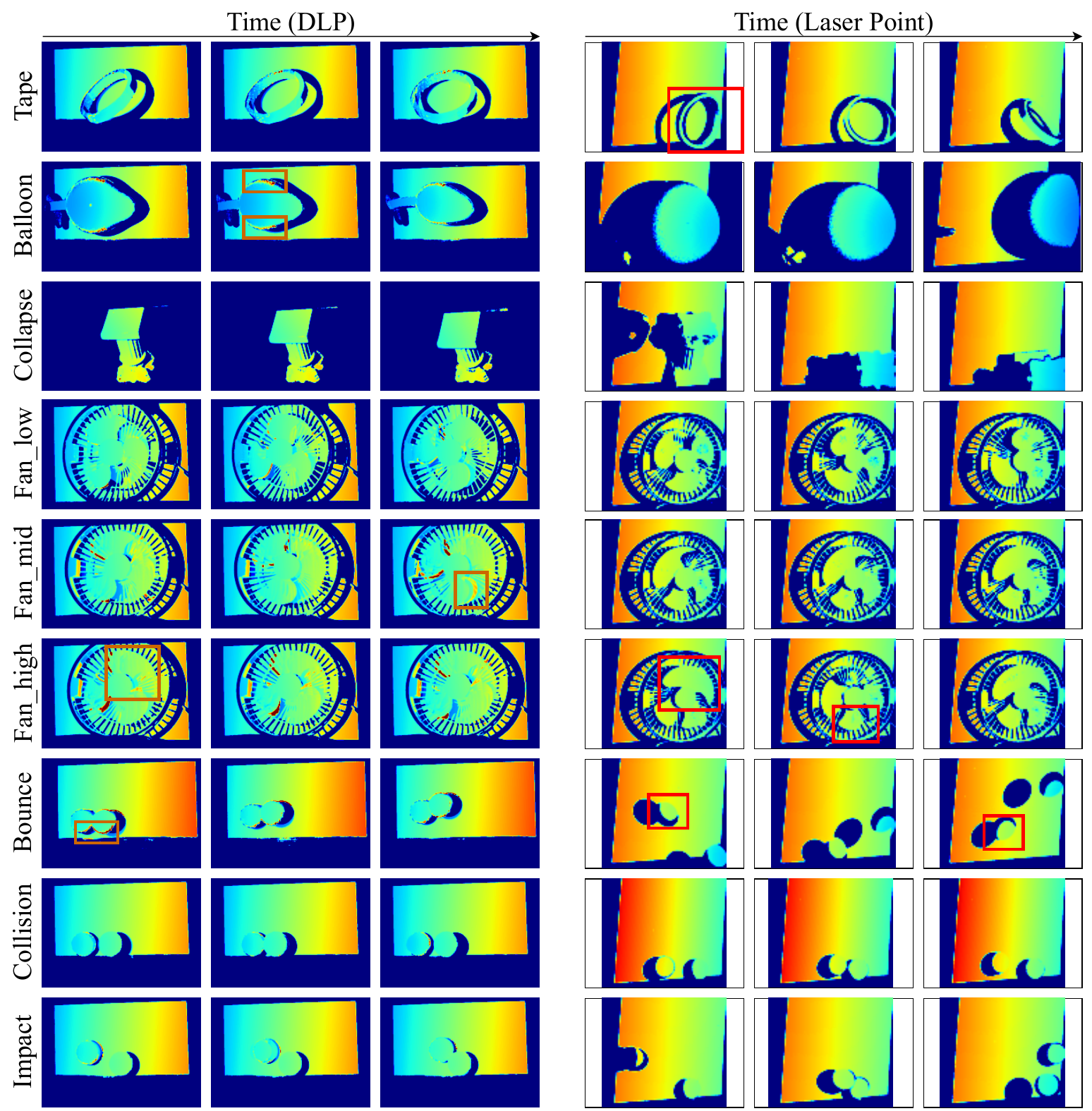}
    \caption{\textbf{Visual results of point scanning methods compared with our SGE on dynamic scenes.} Various dynamic scenes at diverse timings are presented, proving our capacity of depth estimation on dynamic scenes. Given the similar qualitative performance across all point scanning methods, we use the X-map method as a representative. The areas marked by red boxes highlight the rolling shutter effect experienced with point scanning schemes, and those marked by brown boxes highlight the global shutter effect experienced with our SGE. \textbf{Visualization 1-4} in certain dynamic scenes are also provided.}
    \label{fig:dynamic}
    \vspace{-1.5em}
\end{figure*}

We also test the dynamic scene depth estimation capability of our SGE using the DLP-based system with a scanning speed of 402$\mu s$ per slide and compare our results with point scanning methods on the same dynamic scenes, as shown in Fig.~\ref{fig:dynamic}. 
Side-by-side comparison videos showcasing continuous depth estimation in certain dynamic scenes are also provided in \textbf{Visualization 1-4}. Both methods are slowed down by a factor of 60 in these videos.
It is worth noting that our SGE and point scanning methods exhibit completely different characteristics in depth estimation for dynamic scenes, similar to the distinction between global shutter and rolling shutter in traditional frame based cameras. 

We can observe that point-scanning approaches yield good results on low-speed dynamic scenes. However, for high-speed dynamic scenes, although they still produce relatively sharp edges, they suffer from the \textit{\textbf{rolling shutter effect}}. In the sequence Tape, it is easy to observe that a three-dimensional rotating tape becomes twisted; Each blade of a rotating fan has a different size (\eg Fan); Spheres become ellipsoids and the shape of the balls change during the falling and rebounding process in the sequence bounce. The fundamental cause of this effect is that the point-by-point scanning speed of point scanning schemes is \textit{\textbf{lower}} than the speed of the object. Thus, increasing the scanning speed is the fundamental way to reduce the rolling shutter effect. However, point scanning schemes suffer from timestamp noise and low encoding efficiency, which hinders them from higher-speed depth data acquisition.

Although when the speed of dynamic scenes exceeds its capacity, our SGE is also affected by the "global shutter effect", which causes blur and noise around the edges of scenes. For example, as shown in Fig.~\ref{fig:dynamic}, there is blur around the fan blade (\eg Impact) and noise at the edges of the balloon (\eg Balloon) and balls (\eg Bounce). However, since our approach offers \textit{\textbf{higher encoding efficiency}} and \textit{\textbf{immunity to timestamp noise}}, we can realize higher depth data acquisition speed, thereby mitigating the impact of motion. This enables us to perform depth estimation for more rapid scenes which is unachievable by other point scanning methods. Our SGE has achieved satisfactory results in various challenging high-speed dynamic scenarios, such as multi-object interaction (\eg Bounce, Collision, and Impact), thin structures at different speeds (\eg Fan), and three-dimensional rotation (\eg Tape). This proves that our SGE is capable of high-speed depth estimation, not only in simple scenes but also in various motion modes and complex scenes.

Since it is difficult to obtain recognized high-precision depths of dynamic scenes at $\mu s$ temporal resolution without relying on high-speed cameras or other costly equipment, we are unable to directly prove the accuracy of SGE in such scenarios.
But as the experiment shown in  Sec.~\ref{sec:speed} proves the high accuracy of our SGE at high scanning speeds, it can be assured that the accuracy of our scheme in dynamic scenes is guaranteed. The only additional factor impacting the accuracy is the incorrect depth encoding caused by motion, as the noise and blur around the edge shown in Fig.~\ref{fig:dynamic}. This is a problem that every dynamic scene depth estimation method will encounter. The fundamental solution is to enhance data acquisition, which is the greatest strength of our SGE.

\subsection{Runtime evaluation for real-time processing}
\label{sec:realtime}

\begin{table*}[t]
        \caption{\textbf{Disparity calculation time (in seconds) on static scenes using our GX-map and traditional epipolar line search.}}
        \label{tab:ins}
	\centering
        \small
        \resizebox{1.0\textwidth}{!}{
                \setlength{\tabcolsep}{6pt}
                \renewcommand\arraystretch{1.0}
    	    \begin{tabular}{rcccccccccccccccccccccc}
    		\bottomrule[0.15em]
    		\multirow{1}*{Scene} & \multicolumn{1}{c}{David} & \multicolumn{1}{c}{Dragon} 
                & \multicolumn{1}{c}{Structure} & \multicolumn{1}{c}{gslim}
    		& \multicolumn{1}{c}{boxes}     & \multicolumn{1}{c}{fan}
                & \multicolumn{1}{c}{bottle}    & \multicolumn{1}{c}{wizard}
                & \multicolumn{1}{c}{arena}     & \multicolumn{1}{c}{column}  & \multicolumn{1}{c}{average}
                \\
             
    		\midrule
    		GX-map   & 0.08 & 0.08 & 0.08 & 0.08 & 0.08 & 0.09 
                         & 0.08 & 0.09 & 0.08 & 0.09 & 0.08\\

    		Search   & 18.85 & 17.95 & 22.64 & 21.67 & 20.38 & 17.57  
                      & 17.78 & 19.95 & 19.97 & 19.87 & 19.66 (246$\times$) \\
                      
    		\bottomrule
    	\end{tabular}
        }
\end{table*}

\begin{table*}[t]
        \caption{\textbf{The average total time required for a complete one-time depth estimation in dynamic scenes using both CPU and GPU (measured in seconds) of our SGE.}  
        Note that the maximum speed of other point scanning schemes is constrained by the point scanning speed of 60 Hz.}
        \label{tab:cg}
	\centering
        \small
        \resizebox{1.0\textwidth}{!}{
                \setlength{\tabcolsep}{6pt}
                \renewcommand\arraystretch{1.0}
    	    \begin{tabular}{rcccccccccccccccccccc}
    		\bottomrule[0.15em]
    		\multirow{1}*{Scene} & \multicolumn{1}{c}{Tape} & \multicolumn{1}{c}{Balloon} 
                & \multicolumn{1}{c}{Collapse} & \multicolumn{1}{c}{Fan\_low}
    		& \multicolumn{1}{c}{Fan\_mid}     & \multicolumn{1}{c}{Fan\_high}
                & \multicolumn{1}{c}{Bounce}    & \multicolumn{1}{c}{Collision}
                & \multicolumn{1}{c}{Impact}    & \multicolumn{1}{c}{average}
                \\
             
    		\midrule
    		CPU   & 0.024 & 0.024 & 0.018 & 0.027 & 0.025 & 0.024 
                         & 0.024 & 0.024 & 0.024 & 0.024 (42 Hz)  \\
                         
    		GPU   & 0.005 & 0.005 & 0.004 & 0.005 & 0.005 & 0.005  
                      & 0.005 & 0.005 & 0.006 & 0.005 (200 Hz)\\
                      
    		\bottomrule
    	\end{tabular}
        }
\end{table*}

Additionally, we evaluate the computational cost of our SGE to illustrate the ability of our method to realize dense and high-precision depth estimation in real time. We use Python 3.9 on 12th Gen Intel i7-12700F CPU and RTX 3060 GPU to perform these experiments. 

Tab.~\ref{tab:ins} shows the increase in speed using our GX-map disparity query. We evaluate the disparity calculation speed on static scenes using our GX-map and traditional disparity searching in CPU, the latter is the same algorithm ESL used. Two schemes use the same data as input and do not apply any algorithm optimization. It can be seen that our GX-map achieves around 250 times improvement in computational speed.

Tab.~\ref{tab:cg} shows the average total time of once depth map calculation. After simple algorithm optimization, our SGE realizes 200 Hz in GPU and 42 Hz in CPU for real-time depth estimation of 720 * 1280 resolution. It is worth mentioning that the X-map method, which has a similar computational complexity as ours, can also realize real-time processing, but the data acquisition speed of 60 Hz makes it impossible for them to perform faster real-time depth computing.


\section{Conclusion, Limitation and Future Work}
In this work, we propose a high-speed event-based structured light depth estimation scheme: SGE. 
For the first time we introduce Gray code into event-based SL systems for depth estimation with a readily available DLP projector, realizing high depth acquisition speed while maintaining high accuracy.
In addition, due to the absence of a simple and reliable calibration scheme, we present a sub-pixel level, universally applicable calibration method SEC for event-based SL systems.
Furthermore, we introduce the GX-map disparity query scheme and the Time-overlapping strategy to enhance computational speed and data utilization.
Experimental results show that our SGE achieves over kilohertz-level data acquisition speed and around 200 Hz real-time processing speed while maintaining comparable accuracy as the state-of-the-art point scanning methods. 
This proves the capability of our SGE to realize high-speed, high-accuracy, and real-time depth estimation.

Our method is primarily limited by the combined influence of the event camera sensor and illumination conditions, including ambient light and projected light intensity. While using a high projection light intensity and a high event camera threshold can almost entirely mitigate the effects of ambient light, depth-related uneven light intensity distribution, and motion events, this drastically slows the response speed of event cameras and lengthens the refractory period. Therefore, with current devices, we need to comprehensively adjust the ambient light intensity and projection light intensity at a lower event camera threshold to achieve the highest recovery accuracy. However, inappropriate light-intensity conditions may only cause a slight decrease in accuracy. 
Secondly, too fast motion in dynamic leads to depth estimation errors, and these errors will vary when updating the Gray code with different sequences using the Time-overlapping strategy. This can be mitigated by speeding up depth data collection using more advanced hardware. 
Finally, we believe that deeper optimization of the overall algorithm, the utilization of more advanced computing devices, and the substitution of more efficient programming languages can further enhance our real-time computing speed.

\begin{backmatter}
\bmsection{Funding}National Natural Science Foundation of China (12174341) and National Key R\&D Program of China (2022YFF0705500).
\bmsection{Disclosures}The authors declare no conflicts of interest.
\bmsection{Data availability}Data underlying the results presented in this paper are not publicly available at this time but may be obtained from the authors upon reasonable request.
\bmsection{Supplemental document}See Supplement 1 for supporting content and Datasets 1, 2 for calibrated data of the laser-point based and the DLP based system.
\end{backmatter}
\bibliography{main}






\end{document}